\documentclass[twocolumn,conference, letterpaper]{ieeeconf}
\usepackage[T1]{fontenc}
\usepackage[latin9]{inputenc}
\usepackage{xcolor}
\usepackage{cprotect}
\usepackage{calc}
\usepackage{url}
\usepackage{amsmath}
\usepackage{amssymb}
\usepackage{graphicx}
\usepackage[bookmarks=true,bookmarksnumbered=true,bookmarksopen=true,bookmarksopenlevel=1,
 breaklinks=false,pdfborder={0 0 0},pdfborderstyle={},backref=false,colorlinks=false]
 {hyperref}
\hypersetup{pdftitle={UMIRobot: An Open-\{Software, Hardware\} Low-Cost Robotic Manipulator
for Education},
 pdfauthor={Murilo M. Marinho},
 pdfpagelayout=OneColumn, pdfnewwindow=true, pdfstartview=XYZ, plainpages=false}

\makeatletter

\providecommand{\tabularnewline}{\\}

\usepackage[caption=false,font=footnotesize]{subfig}
\usepackage{xcolor}
\usepackage{graphicx}
\usepackage{cite}

\usepackage{resizegather}

\IEEEoverridecommandlockouts 

\makeatother

\usepackage{listings}
\begin{document}
\global\long\def\quat#1{\boldsymbol{#1}}%

\global\long\def\dq#1{\underline{\boldsymbol{#1}}}%

\global\long\def\hp{\mathbb{H}_{p}}%

\global\long\def\dotmul#1#2{\langle#1,#2\rangle}%

\global\long\def\partialfrac#1#2{\frac{\partial\left(#1\right)}{\partial#2}}%

\global\long\def\totalderivative#1#2{\frac{d}{d#2}\left(#1\right)}%

\global\long\def\mymatrix#1{\boldsymbol{#1}}%

\global\long\def\vecthree#1{\operatorname{vec}_{3}#1}%

\global\long\def\vecfour#1{\operatorname{vec}_{4}#1}%

\global\long\def\haminuseight#1{\overset{-}{\mymatrix H}_{8}\left(#1\right)}%

\global\long\def\hapluseight#1{\overset{+}{\mymatrix H}_{8}\left(#1\right)}%

\global\long\def\haminus#1{\overset{-}{\mymatrix H}_{4}\left(#1\right)}%

\global\long\def\haplus#1{\overset{+}{\mymatrix H}_{4}\left(#1\right)}%

\global\long\def\norm#1{\left\Vert #1\right\Vert }%

\global\long\def\abs#1{\left|#1\right|}%

\global\long\def\conj#1{#1^{*}}%

\global\long\def\veceight#1{\operatorname{vec}_{8}#1}%

\global\long\def\myvec#1{\boldsymbol{#1}}%

\global\long\def\imi{\hat{\imath}}%

\global\long\def\imj{\hat{\jmath}}%

\global\long\def\imk{\hat{k}}%

\global\long\def\dual{\varepsilon}%

\global\long\def\getp#1{\operatorname{\mathcal{P}}\left(#1\right)}%

\global\long\def\getpdot#1{\operatorname{\dot{\mathcal{P}}}\left(#1\right)}%

\global\long\def\getd#1{\operatorname{\mathcal{D}}\left(#1\right)}%

\global\long\def\getddot#1{\operatorname{\dot{\mathcal{D}}}\left(#1\right)}%

\global\long\def\real#1{\operatorname{\mathrm{Re}}\left(#1\right)}%

\global\long\def\imag#1{\operatorname{\mathrm{Im}}\left(#1\right)}%

\global\long\def\spin{\text{Spin}(3)}%

\global\long\def\spinr{\text{Spin}(3){\ltimes}\mathbb{R}^{3}}%

\global\long\def\distance#1#2#3{d_{#1,\mathrm{#2}}^{#3}}%

\global\long\def\distancejacobian#1#2#3{\boldsymbol{J}_{#1,#2}^{#3}}%

\global\long\def\distancegain#1#2#3{\eta_{#1,#2}^{#3}}%

\global\long\def\distanceerror#1#2#3{\tilde{d}_{#1,#2}^{#3}}%

\global\long\def\dotdistance#1#2#3{\dot{d}_{#1,#2}^{#3}}%

\global\long\def\distanceorigin#1{d_{#1}}%

\global\long\def\dotdistanceorigin#1{\dot{d}_{#1}}%

\global\long\def\squaredistance#1#2#3{D_{#1,#2}^{#3}}%

\global\long\def\dotsquaredistance#1#2#3{\dot{D}_{#1,#2}^{#3}}%

\global\long\def\squaredistanceerror#1#2#3{\tilde{D}_{#1,#2}^{#3}}%

\global\long\def\squaredistanceorigin#1{D_{#1}}%

\global\long\def\dotsquaredistanceorigin#1{\dot{D}_{#1}}%

\global\long\def\crossmatrix#1{\overline{\mymatrix S}\left(#1\right)}%

\global\long\def\constraint#1#2#3{\mathcal{C}_{\mathrm{#1},\mathrm{#2}}^{\mathrm{#3}}}%

\title{UMIRobot: An Open-\{Software, Hardware\} Low-Cost Robotic Manipulator
for Education}
\author{Murilo~M.~Marinho, Hung-Ching Lin$^{*}$, Jiawei Zhao$^{*}$\thanks{$^{*}$The
authors H. Lin and J. Zhao had equal contributions, described in Sections
\ref{subsec:quentin} and \ref{subsec:jzhao}, respectively.}\thanks{This
work was supported in part by the Institute for Innovation in International
Engineering Education of the School of Engineering, the University
of Tokyo, and in part by JSPS KAKENHI Grant Number JP19K14935. }\thanks{(\emph{Corresponding
author:} Murilo~M.~Marinho)}\thanks{Murilo M. Marinho, Hung-Ching
Lin, and Jiawei Zhao are with the Department of Mechanical Engineering,
the University of Tokyo, Tokyo, Japan. \texttt{Emails:}\{murilo, qlin1806,zhao-jiawei336\}@g.ecc.u-tokyo.ac.jp.
}}
\maketitle
\begin{abstract}
Robot teleoperation has been studied for the past 70 years and is
relevant in many contexts, such as in the handling of hazardous materials
and telesurgery. The COVID19 pandemic has rekindled interest in this
topic, but the existing robotic education kits fall short of being
suitable for teleoperated robotic manipulator learning. In addition,
the global restrictions of motion motivated large investments in online/hybrid
education. In this work, a newly developed robotics education kit
and its ecosystem are presented which is used as the backbone of an
online/hybrid course in teleoperated robots. The students are divided
into teams. Each team designs, fabricates (3D printing and assembling),
and implements a control strategy for a master device and gripper.
Coupling those with the UMIRobot, provided as a kit, the students
compete in a teleoperation challenge. The kit is low cost (< 100USD),
which allows higher-learning institutions to provide one kit per student
and they can learn in a risk-free environment. As of now, 73 such
kits have been assembled and sent to course participants in eight
countries. As major success stories, we show an example of gripper
and master designed for the proposed course. In addition, we show
a teleoperated task between Japan and Bangladesh executed by course
participants. Design files, videos, source code, and more information
are available at \url{https://mmmarinho.github.io/UMIRobot/}
\end{abstract}

\section{Introduction}

Robot teleoperation was born in the 1940s, in the context of handling
radioactive materials \cite{Hokayem_2006}. In essence, teleoperation
means that a \emph{leader} robot is used to control a \emph{follower}
robot though a communication channel. In contrast with robotic applications
that allow offline planning, teleoperation in general requires a fast
response time as it is inherently interactive.

The first teleoperated system featured a mechanically coupled leader-follower
pair robot and, after over 50 years of development in myriad fields
of science and technology, we saw the first example of robot teleoperation
over the internet \cite{Goldberg}. In a relatively short period,
teleoperation technologies enabled a transatlantic robot-assisted
telesurgical procedure \cite{Marescaux_2001}. After a couple of decades,
as of 2022, the teleoperated da Vinci Surgical System (Intuitive Surgical,
USA) family of systems has been used to perform over 10 million surgical
procedures\footnote{Intuitive Surgery: .P. Morgan Healthcare Conference 2022 Presentation.
\url{https://isrg.gcs-web.com/static-files/e649a656-539c-4e52-8445-4d6135e06658}}. In addition, the natural isolation of medical staff and patients
provided by teleoperation has also motivated the development of teleoperated
robots in COVID-19 contexts \cite{Chen_2022}.

Given the constant evolution of those robotic systems, many groups
have identified that the costs of state-of-the-art manipulators with
multiple degrees of freedom (DoFs) are prohibitive for many research
groups \cite{Kazanzides_2014,Wiedmeyer_2019}. This has been addressed
in part by the development of open-\{software, hardware\} systems
with donated hardware \cite{Kazanzides_2014} and internet-accessible
robot experiment platforms \cite{Takacs_2016,Wiedmeyer_2019}. This
way, the same robotic system can be shared by a larger number of researchers
and students, helping to dissolve the costs of purchasing and maintaining
those systems.

\begin{figure}[t]
\begin{centering}
\includegraphics[width=1\linewidth]{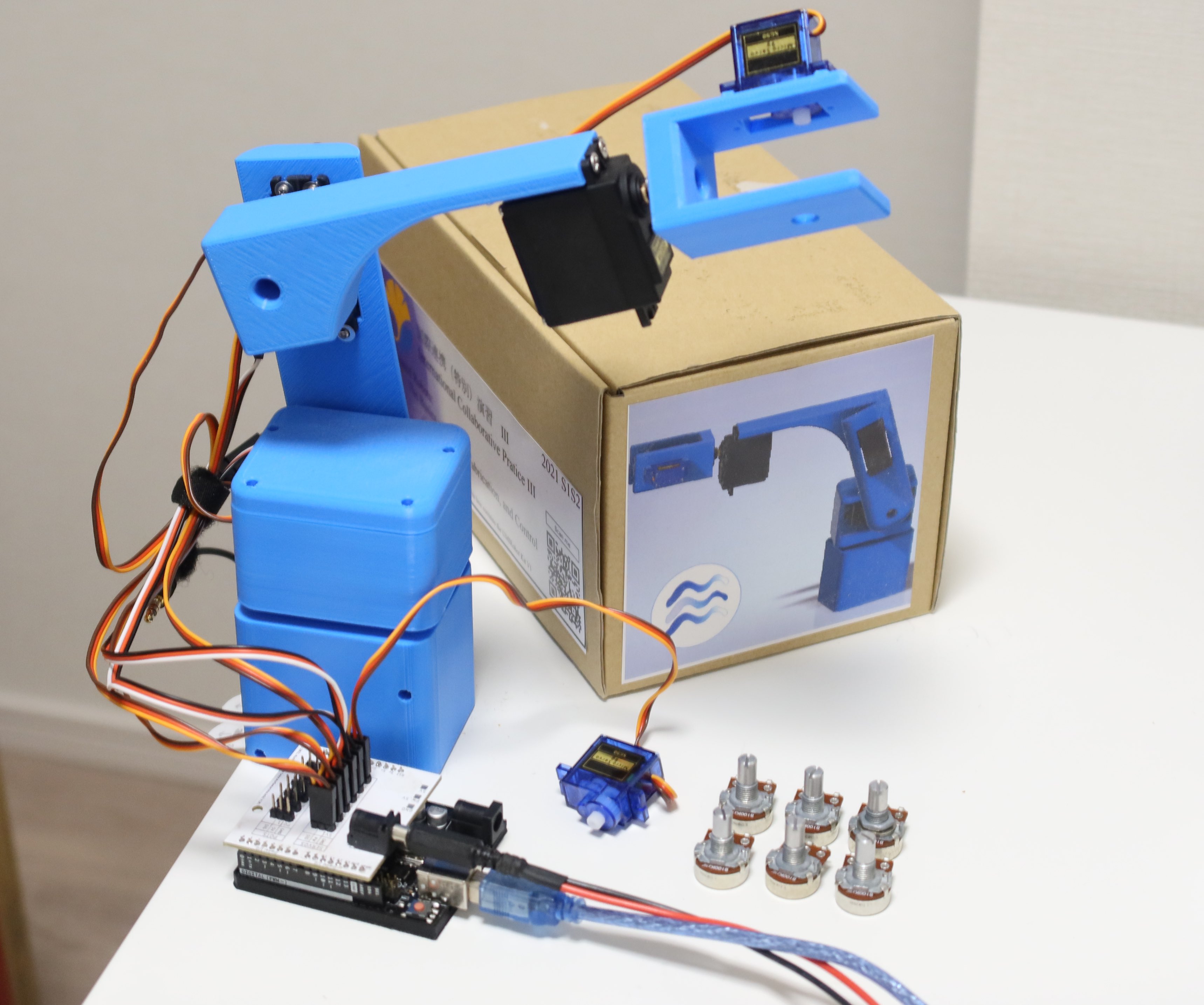}
\par\end{centering}
\caption{\protect\label{fig:realumirobot} The fully assembled UMIRobot, other
important elements of the kit, and the box used to deliver the kit.}
\end{figure}
Nonetheless, we have identified a gap in robot learning kits for courses
in (under)graduate education in manipulator robot teleoperation. An
effect of this education gap is that most students joining research
groups have little to no practice in handling manipulator robots.
When certain students have the opportunity to use them, the students
have limited freedom in controlling those robots because institutions
cannot risk hurting students, or personnel, or breaking robots. Having
a low-cost \emph{individual} robot kit that each student can keep
after the end of the course removes this psychological pressure and
the student can focus on learning in an environment with no financial
risk.

Moreover, the restriction of motion imposed by the ongoing COVID-19
pandemic has fueled an unprecedented expansion of online learning.
At the School of Engineering of the University of Tokyo, similarly
to other higher-learning institutions worldwide, in-person learning
was limited and online learning was incentivized. In this context,
students needed a low-cost (<100 USD per student) platform that they
could use at their own homes. In addition, a large number of international
students were unable to enter their country of study given border
control measures. The low-cost nature of the kit serves three main
purposes. First, because the kits are entirely funded by the University,
we can reach a larger number of students within a set budget. Second,
it allows the students to receive the learning kit through mail and
not be required to return it. Third, it facilitates customs clearance,
as many countries have an upper limit of approximately 100 USD for
the simplified imports of goods for noncommercial use, whereas it
would be administratively unmanageable for the University to pay any
extra import fees on behalf of students.

As the world returns to normalcy and students go back to in-person
learning, an added benefit of having such a platform is that students
enrolled in universities in other countries can also participate and
improve the globalization of higher learning institutions.

\subsection{Related Works}

Most existing educational platforms do not target education with a
robotic manipulator \cite{Ma_2013,Casan_2015,Yu_2017,Paull_2017,Brand_2018,Lin_2019,Ramos_2021},
despite robotic manipulators being ubiquitous in the manufacturing/medical
industry and robotics research. For instance, Ma et al. \cite{Ma_2013}
developed an open-source 3D printed underactuated hand. Casan et al.
\cite{Casan_2015} proposed an online robot programming framework
for remote education including mobile and humanoid robots. Yu et al.
\cite{Yu_2017} developed a portable 3D printed multi-vehicle platform.
Duckietown \cite{Paull_2017} and Duckiepond \cite{Lin_2019} are
open educational environments for driving automation and maritime
automation, respectively. Brand et al. \cite{Brand_2018} explored
drone education with a platform based on Raspberry Pi and Python.

Other research/educational platforms contain a manipulator but are
not low-cost enough for exportation and to allow the students to keep
the hardware \cite{Bischoff_2011,Kazanzides_2014}. Bischoff et al.
\cite{Bischoff_2011} proposed the KUKA youBot mobile manipulator
with impressive capabilities and widely used worldwide, but with price
and size incompatible with the targets of this work. Lastly, Kazanzides
et al. \cite{Kazanzides_2014} proposed the well-known da Vinci Research
Kit, widely used in surgical robotics research. The hardware is based
on donated decommissioned surgical robots, but those can only be given
to research institutions and not individuals.

Note that even though many research institutes have relied on commercial
education kits \cite{Takacs_2016} such as the LEGO Mindstorm NXT
\cite{Ruzzenente12areview} (currently discontinued), none of the
existing systems provide a complete 5-DoFs or more manipulator robot
kit in the necessary price/size range that allows each student to
receive the kits anywhere in the world and keep their kit after the
the course is over. The form factor affects shipping costs which are included
in the total price of goods for customs clearance, hence a larger
form factor can make a considerable difference in the feasibility
of such kit.

\subsection{Statement of Contributions}

In this work, we describe the UMIRobot and its learning ecosystem.
The contributions of this work are
\begin{enumerate}
\item a miniature manipulator robot design that can be effectively 3D printed,
called UMIRobot. It has five joint DoF and one additional DoF for
a gripper. All design files are available at Thingiverse.\footnote{\url{https://www.thingiverse.com/thing:4797804}}
The robot can be packed into a low-cost educational kit (< 100USD).
Shown in Section~\eqref{subsec:Proposed-miniature-manipulator}.
\item an open-source printed circuit board (PCB) design to easily interface
the robot (and a possible master interface) to Arduino UNO and compatible
designs, such as Arduino Leonardo. Shown in Section~\eqref{subsec:Electrical}.
\item an open-source Arduino library available in the Arduino Library Manager
used to process the information from the UMIRobot.\footnote{\url{https://github.com/mmmarinho/umirobot-arduino}}
Shown in Section~\eqref{subsec:Programming}.
\item an open-source Python3 library to read from the UMIRobot and control
it.\footnote{\url{https://github.com/mmmarinho/umirobot-py}} The
library has an intuitive graphical user interface (GUI) targeting
early undergraduate students. Shown in Section~\eqref{subsec:Programming}.
\item several open-source examples in Python3 that serve as educational
examples for the teleoperation of the UMIRobot.\footnote{\url{https://github.com/mmmarinho/umirobot}}
We also shown in this work two successful examples to showcase the
capabilities of this work in Section~\eqref{sec:Achievement-showcase}.
\item a two-year-long analysis of the student corpus and feedback. Shown
in Section~\eqref{sec:Student-distribution-and}.
\end{enumerate}

\section{Requirements}

The course at the University of Tokyo that generated the demand for
the robot kit is entitled ``Teleoperated robots: the basics of design,
fabrication, and control''. The students organize themselves into
groups and work on designing a gripper and a master device. The students
must be able to receive the kits internationally and cooperate with
their partners abroad. In that sense, we had the following requirements
\begin{enumerate}
\item the robot and its main components can be assembled with only one screwdriver,
e.g. does not require soldering,
\item the students must be able to move the robot from their own personal
computers,
\item the entire kit must fit in the size and weight restrictions for international
shipping, e.g. DHL's smallest\footnote{Currently at 337 x 182 x 100 mm (1 kg).}
box,
\item the kit must not contain batteries and other commonly prohibited items
in shipping that would further complicate logistics.
\end{enumerate}
Owing to administrative/budget procedures, we could only one kit to
each student. However, in special for students participating from
abroad, they had to manage any repair or replacement parts (if needed).
Hence, having parts easy to print or procure was an important requirement.

\section{Proposed educational kit\protect\label{subsec:RobotKit}}

\begin{figure}[t]
\begin{centering}
\noindent\fcolorbox{black}{white}{\begin{minipage}[t]{1\columnwidth - 2\fboxsep - 2\fboxrule}%
\begin{center}
\includegraphics[width=1\columnwidth]{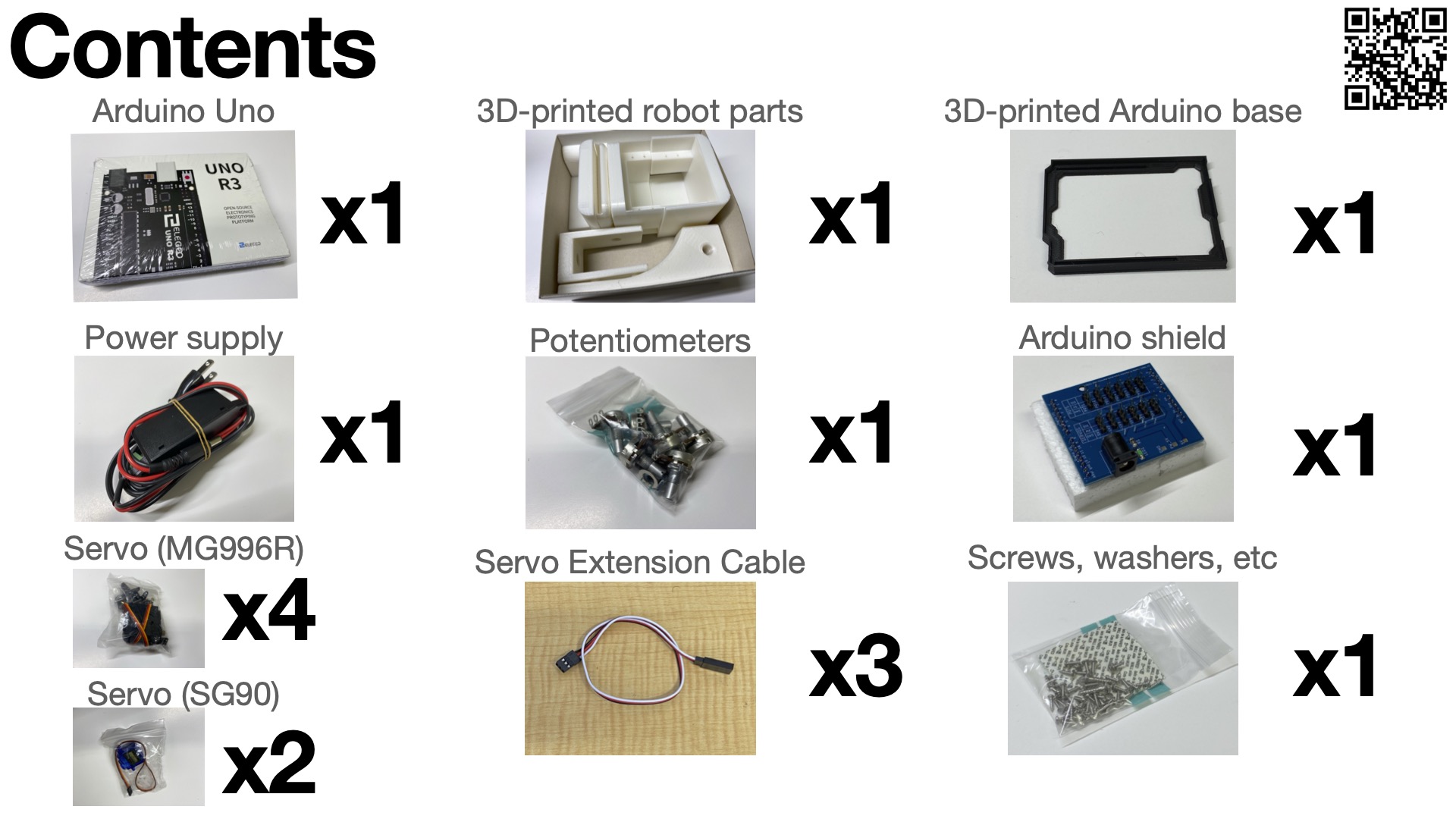}
\par\end{center}%
\end{minipage}}
\par\end{centering}
\caption{\protect\label{fig:kit_contents} All elements contained in the kit.
This listed is included in the kit so that each student can check
if the kit is complete.}
\end{figure}

The contents of the robot kit are summarized in Table~\ref{tab:Estimated-costs-of},
among with the total estimated cost. The contents of the kit are also
shown in Fig. ~\ref{fig:kit_contents}.

\begin{table}[tbh]
\caption{\protect\label{tab:Estimated-costs-of}Estimated costs of one UMIRobot
kit.}

\medskip{}

\centering{}%
\begin{tabular}{|c|c|c|c|}
\hline 
Content & Quantity & Group & Cost\tabularnewline
\hline 
\hline 
3D Printer Filament & 146 g & Mechanical & \textasciitilde 5 USD\tabularnewline
\hline 
Arduino UNO & 1 & Electronic & \textasciitilde 13 USD\tabularnewline
\hline 
Custom PCB & 1 & Electronic & \textasciitilde 5 USD\tabularnewline
\hline 
Potentiometers & 6 & Electronic & \textasciitilde 5 USD\tabularnewline
\hline 
IRM-30-5ST & 1 & Power & \textasciitilde 20 USD\tabularnewline
\hline 
MG996R & 4 & Electro-Mechanic & \textasciitilde 10 USD\tabularnewline
\hline 
SV90 & 2 & Electro-Mechanic & \textasciitilde 3 USD\tabularnewline
\hline 
Screws, Cables, ... &  &  & \textasciitilde 9 USD\tabularnewline
\hline 
\hline 
Total &  &  & \textbf{\textasciitilde 70 USD}\tabularnewline
\hline 
\end{tabular}
\end{table}

The kit is mostly composed by components that are popular with hobbyists,
making them easy to procure as they are produced in large scale and
available worldwide\cprotect\footnote{On the course website, we mention where to procure the parts in Japan.
However, these parts can be found in sites available worldwide such
as \url{alibaba.com}, \url{ebay.com}, \url{amazon.your_country},
\url{digikey.com} and so on.}.

\subsection{Proposed miniature manipulator design\protect\label{subsec:Proposed-miniature-manipulator}}

\begin{figure}[t]
\begin{centering}
\includegraphics[width=1\linewidth]{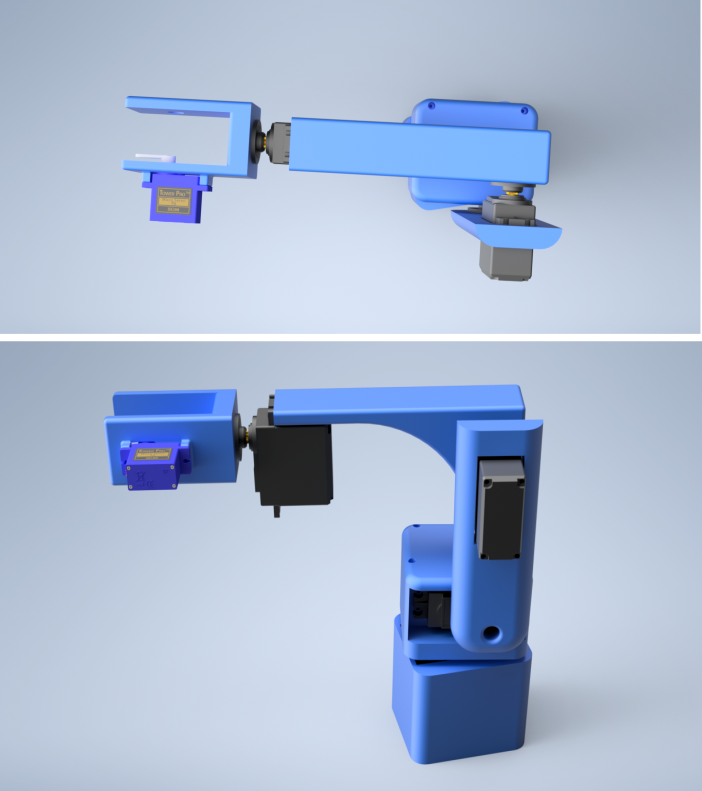}
\par\end{centering}
\caption{CAD rendering of the assembled 5 DoF UMIRobot, the miniature manipulator
robot newly designed in this work. Note that the fifth joint, $q_{4}$,
will be attached to the student's gripper module to rotate it. A sixth
servomotor (not shown in the CAD) will be used to actuate the gripper.
\textbf{Design files available at \protect\url{https://www.thingiverse.com/thing:4797804}.}}
\end{figure}

The UMIRobot is a novel design aiming for this online course. The body
of the robot was designed to be printed with a regular FDM printer.
With printers normally available at universities (X-MAX, Qidi Technology),
the entire robot can be printed at once. The CAD design of the robot
is shared at Thingiverse\footnote{\url{https://www.thingiverse.com/thing:4797804}}
licensed under CC BY-NC-ND 4.0. The whole robot can be assembled with
a single screwdriver. Instructions on how to assemble the robot are
part of the course and available in the course's main website.

\subsection{Kinematic modeling}

\begin{table}[t]
\caption{\protect\label{tab:DH-parameters-of}DH parameters of the UMIRobot.}

\begin{center}

\noindent\resizebox{1.0\columnwidth}{!}{%

\begin{tabular}{cccccc}
\hline 
 & 1 & 2 & 3 & 4 & 5\tabularnewline
\hline 
\hline 
$\theta_{\text{DH}}$ {[}rad{]} & 0 & $-\pi/2$ & 0 & $0$ & 0\tabularnewline
$d_{\text{DH}}$ {[}m{]} & $0.00245$ & $0$ & $0$ & $0.16519$ & $0$\tabularnewline
$a_{\text{DH}}$ {[}m{]} & $0$ & $0.0813$ & $0$ & $0$ & $0$\tabularnewline
$\alpha_{\text{DH}}$ {[}rad{]} & $-\pi/2$ & $\pi$ & $\pi/2$ & $-\pi/2$ & $\pi/2$\tabularnewline
\hline 
\end{tabular}

}

\end{center}
\end{table}

The UMIRobot is composed of five revolute joints in series, namely
$\myvec q=\begin{bmatrix}q_{0} & q_{1} & q_{2} & q_{3} & q_{4}\end{bmatrix}^{T}\in\mathbb{R}^{5}$.
The last joint, $q_{4}$, is used to rotate the grippers designed
by the students. Hence, the robot has five DoF. It can be modelled
using Denavit\textendash Hartenberg (DH) parameters as shown in Table~\ref{tab:DH-parameters-of}.
Each joint is limited by $\pm\pi/2$ rad.

The sixth and last servomotor is used to open and close the gripper.
Accordingly, the sixth servomotor is not part of the kinematic model.

\subsection{Kinematic control\protect\label{subsec:Kinematic-control}}

The UMIRobot can be controlled in two modes, enabled by the software
described in Section~\ref{subsec:Programming}. The first mode is
configuration-space control in which the user defines a target joint
position $\myvec q_{d}\in\mathbb{R}^{5}$ and sends it to the robot.
The second mode is a state-of-the-art constrained kinematic control\footnote{Implemented for the UMIRobot in: \url{https://github.com/mmmarinho/umirobot}
using DQRobotics \cite{Adorno2021} in Python3.} (for detailed a course on this topic, refer to \url{https://github.com/dqrobotics/learning-dqrobotics-in-matlab/tree/master/robotic_manipulators})
born in surgical robotics research, given by \cite{Marinho2019}
\begin{alignat}{1}
\myvec u\in\underset{\dot{\myvec q}}{\text{arg min}}\  & \alpha\norm{\mymatrix J_{t}\dot{\myvec q}+\eta\vecfour{\tilde{\quat t}}}_{2}^{2}\label{eq:cost_function_translation}\\
 & +\left(1-\alpha\right)\norm{\mymatrix J_{r}\dot{\myvec q}+\eta\vecfour{\tilde{\myvec r}}}_{2}^{2}\label{eq:cost_function_rotation}\\
 & +\lambda^{2}\norm{\dot{\myvec q}}_{2}^{2}\label{eq:cost_function_joint_damping}\\
\text{subject to}\  & \mymatrix W\dot{\myvec q}\preceq\myvec w,\label{eq:constraints}
\end{alignat}
where $0\leq\alpha\leq1$; the first cost function \eqref{eq:cost_function_translation}
is related to the end-effector translation,\footnote{The translation is written as a pure quaternion in the form $\quat t=t_{x}\imi+t_{y}\imj+t_{z}\imk$
with $t_{x},t_{y},t_{z}\in\mathbb{R}$ and $\imi^{2}=\imj^{2}=\imk^{2}=\imi\imj\imk=-1$.
Moreover, $\vecfour{\quat t}=\begin{bmatrix}0 & t_{x} & t_{y} & t_{z}\end{bmatrix}^{T}\in\mathbb{R}^{4}.$} $\quat t$; the second cost function \eqref{eq:cost_function_rotation}
is related to the end-effector rotation,\footnote{The rotation is written as a unit quaternion in the form $\quat r=\cos\left(\phi/2\right)+\quat v\sin\left(\phi/2\right)$
where $\phi\in\mathbb{R}$ is the angle of rotation about the unit-axis
defined by the pure quaternion $\quat v=v_{x}\imi+v_{y}\imj+v_{z}\imk$.
In addition, $\vecfour{\quat r}=\begin{bmatrix}\cos\left(\phi/2\right) & \sin\left(\phi/2\right)v_{x} & \sin\left(\phi/2\right)v_{y} & \sin\left(\phi/2\right)v_{z}\end{bmatrix}^{T}\in\mathbb{R}^{4}.$
Note that $\quat r^{*}=\cos\left(\phi/2\right)-\quat v\sin\left(\phi/2\right)$
is the conjugate of $\quat r$ and represents the inverse rotation.} $\quat r$; and the third cost function \eqref{eq:cost_function_joint_damping}
is related to damping the joint velocities with damping factor $\lambda\in\left(0,\infty\right)\subset\mathbb{R}$.
In those, we also use the translation Jacobian $\mymatrix J_{\quat t}\in\mathbb{R}^{4\times5}$,
a translation error $\tilde{\quat t}\triangleq\quat t-\quat t_{d}$,
a rotation Jacobian $\mymatrix J_{\quat r}\in\mathbb{R}^{4\times5}$,
and a switching rotational error $\tilde{\quat r}\left(\quat r,\quat r_{d}\right)\triangleq\tilde{\quat r}$
given by
\[
\tilde{\quat r}\triangleq\begin{cases}
\conj{\left(\myvec r\right)}\myvec r_{d}-1 & \text{if }\norm{\conj{\myvec r}\myvec r_{d}-1}_{2}<\norm{\conj{\myvec r}\myvec r_{d}+1}_{2}\\
\conj{\left(\myvec r\right)}\myvec r_{d}+1 & \text{otherwise},
\end{cases}
\]
where $\quat r_{d}$ and $\quat r$ are the desired and current end-effector
rotations, respectively. The desired translation signals, $\quat t_{d}$,
and rotational signals, $\quat r_{d}$, are obtained by the masters
designed by the students. Furthermore, $\eta\in\left(0,\infty\right)\subset\mathbb{R}$
is a proportional gain to reduce the task error. The linear constraints
are used to the robot within its joint limits with
\begin{equation}
\underbrace{\begin{bmatrix}-\mymatrix I_{5\times5}\\
\mymatrix I_{5\times5}
\end{bmatrix}}_{\mymatrix W}\dot{\myvec q}\preceq\underbrace{\begin{bmatrix}\tilde{\myvec q}_{\min}\\
-\tilde{\myvec q}_{\max}
\end{bmatrix}}_{\myvec w},\label{eq:configurations_only_constraints}
\end{equation}
where $\tilde{\myvec q}_{\max}\triangleq\myvec q-\myvec q_{\max}$,
$\tilde{\myvec q}_{\min}\triangleq\myvec q-\myvec q_{\min}$, and
$\mymatrix I_{5\times5}\in\mathbb{R}^{5\times5}$ is the identity
matrix.

At each time step with sampling time $T$, the signal $\myvec u$
in \eqref{eq:cost_function_translation} is computed using a numeric
solver\footnote{E.g. \url{https://github.com/quadprog/quadprog}.},
in less than a millisecond even using a budget laptop. Then, the next
joint position is obtained through simple integration as follows
\[
\myvec q_{d}\left(t+T\right)=\myvec q_{d}\left(t\right)+\myvec uT.
\]
In the example code given to the students, we have $\alpha=0.999$,
$\lambda=0.01$, $\eta=4.0$, and $T$ depends on their computer hardware.
Using a state-of-the-art kinematic control strategy helps the students
generalize to robots with a large number of DoF in future courses.

\subsection{Electrical\protect\label{subsec:Electrical}}

\begin{figure}[t]
\begin{centering}
\includegraphics[width=0.79\columnwidth]{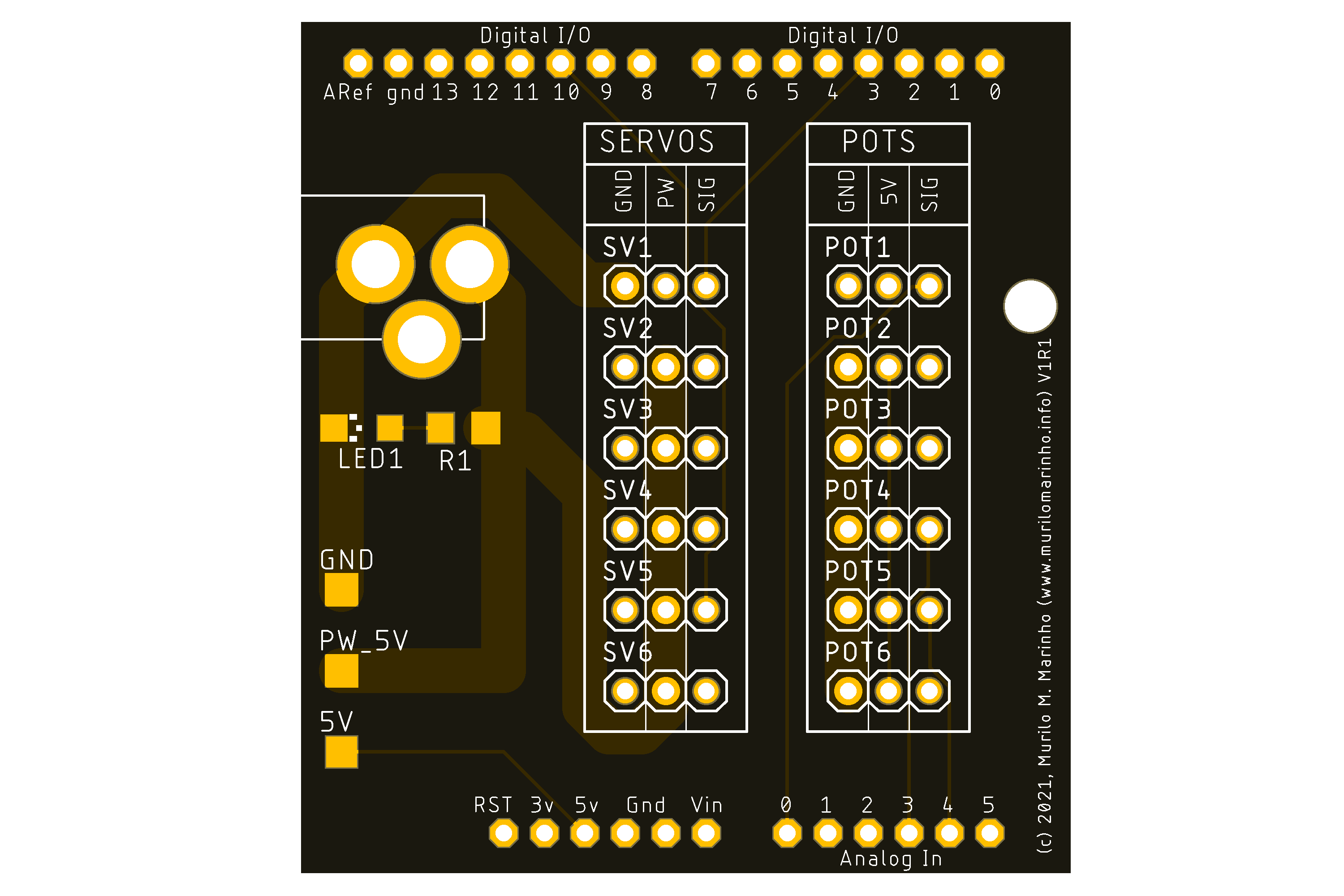}
\par\end{centering}
\caption{\protect\label{fig:pcb}Rendering of the PCB designed as part of the
proposed kit. PCB design and schematics licensed under CC BY-NC-ND
4.0. It has been designed with standards compatible with low-cost
PCB production services. \textbf{Design files available at \protect\url{https://mmmarinho.github.io/UMIRobot/}.}}
\end{figure}

A novel printed circuit board (PCB) is one of the contributions of
this work and provided as part of the kit. The schematics are shown
in Fig. \ref{fig:pcb}.

The PCB was made in such a way to isolate the electrical powering
of the servos from the Arduino. The servos, combined, can use large
amounts of current, and the 5A power source was chosen to provide
for those needs. An LED shows whether the power supply is correctly
connected to help remotely troubleshoot more easily. The servos and
potentiometers can be connected to regular-sized header pins, which
are the default in hobby servos such as the ones used in this kit.
The PCB has several markings so that the students can correctly connect
servos and potentiometers. Lastly, the PCB has been designed to be
used as a shield for the Arduino. The PCBs are provided in their final
assembled form, hence the students can control the robot through their
computers without having to solder.

\subsection{Programming\protect\label{subsec:Programming}}

The microcontroller (Arduino UNO) can be loaded with the necessary
program by installing the GPLv3 UMIRobot library\footnote{\url{https://github.com/mmmarinho/umirobot-arduino.}}
available in the Arduino library manager\footnote{\url{https://www.ardu-badge.com/UMIRobot.}}.
This library is a contribution of this work.

The example sketch, called \texttt{UMIRobot.ino}, must be compiled
and uploaded to the Arduino. Such a process can be done when preparing
the kits, but making it part of the course is more instructive. In
addition, when the course targets students with possibly no prior
experience in programming, it is paramount to have something easy
to install that works out of the box. Gladly, this approach also gives
room to students already accustomed to Arduino programming and gives
them the flexibility to write their own Arduino software.

\begin{figure}[t]
\begin{centering}
\includegraphics[width=1\linewidth]{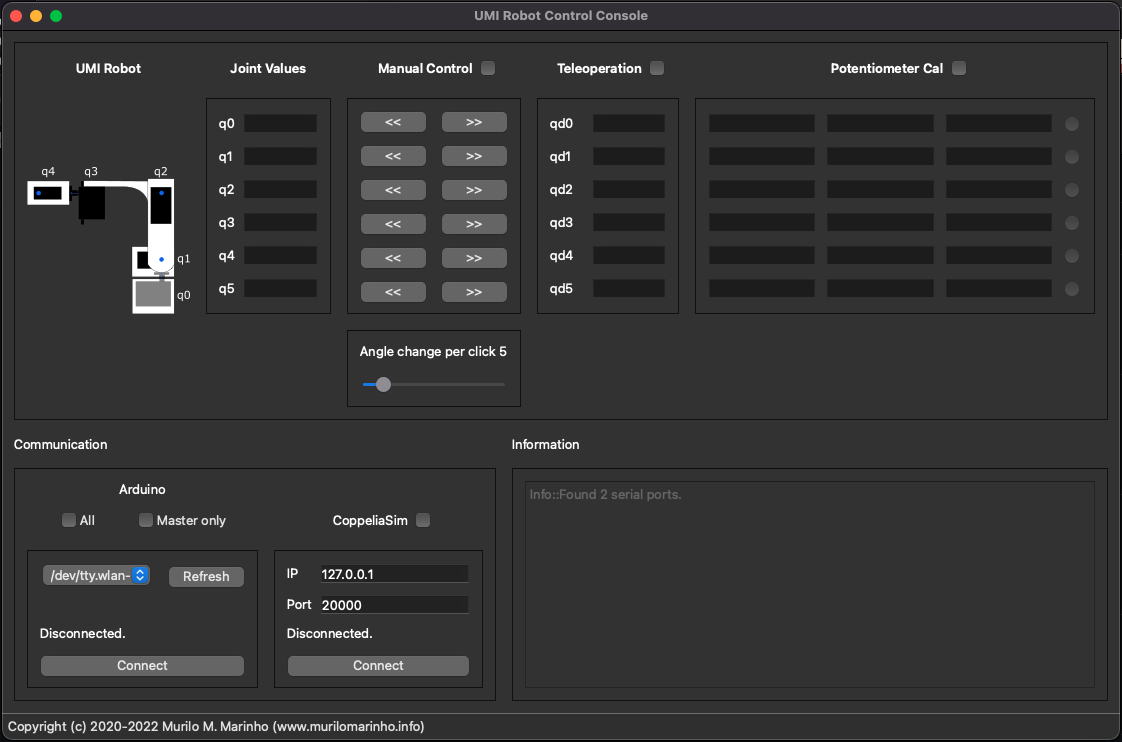}
\par\end{centering}
\caption{\protect\label{fig:gui}A screenshot of the latest version of the
UMIRobot GUI developed in this work. A user with no programming experience
can easily connect to the Arduino and/or the simulator. In addition,
configuration-space control is available out-of-the-box. \textbf{Source
files available at \protect\url{https://github.com/mmmarinho/umirobot-py}.}}
\end{figure}

\subsubsection{Frontend\protect\label{subsec:Frontend}}

In particular for students without any programming or control background,
a graphic user interface (GUI) program\footnote{\url{https://github.com/mmmarinho/umirobot-py}}
has been developed as a contribution of this work. The GPLv3 frontend,
shown in Fig.~\ref{fig:gui}, allows the user to teleoperate the
UMIRobot in several different modes without any prior programming
experience. The user can freely choose to control the real robot or
a simulated version of it by choosing the proper settings in the GUI.
In addition, because the masters designed by the students are based
on potentiometers, there is an automatic potentiometer calibration
mode that obtains the lowest and largest potentiometer values to send
signals within the motion range of each servo. The user can install
the GUI program as a Python package available at Python Package Index
(PyPI)\footnote{\url{https://pypi.org/project/umirobot/}} by running
the following command on a Windows, MacOS, or Linux environment

\begin{lstlisting}
python3 -m pip install umirobot
\end{lstlisting}
and can execute the GUI with the following command

\begin{lstlisting}
python3 -m umirobot
\end{lstlisting}

The students can also download a binary version\footnote{\url{https://github.com/mmmarinho/umirobot-py/latest}}
available for Windows and MacOS.

\subsubsection{Backend\protect\label{subsec:Backend}}

The backend programs, also newly developed in this work, are based
on Python's \texttt{multiprocessing.shared\_memory} module. One process
is used to communicate with the Arduino, obtain its current state
and update its desired state via serial communication. Those data
are stored in the shared memory, which can be accessed by the GUI
program (and other programs). This way, the user can have a smooth
experience.

Students with a stronger background in programming and control can
make customized programs by making use of the backend software. The
most common use for this so far has been for students that design
masters based on task-space control. Different from configuration-space
control, such design has more peculiarities that would be difficult
to provide in a clean GUI. The students are encouraged to develop
their own programs and they can base them on example programs\footnote{\url{https://github.com/mmmarinho/umirobot}}
provided in the course. This is usually reserved for students with
previous knowledge of rigid body motion, but learning resources are
also available\footnote{\url{https://github.com/dqrobotics/learning-dqrobotics-in-matlab}}.
Although learning robotics in-depth in such a short time would be
impossible for most participants, those interested in more complex
control strategies have their needs sufficed.

\subsection{Simulation environment}

\begin{figure}[t]
\begin{centering}
\includegraphics[width=1\linewidth]{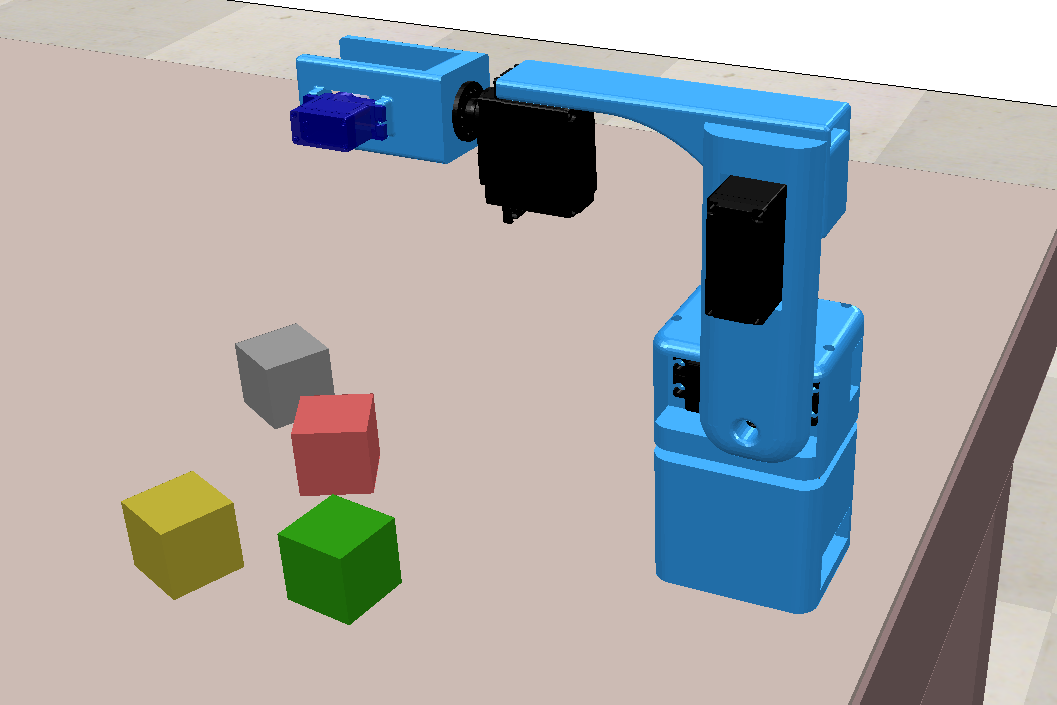}
\par\end{centering}
\caption{\protect\label{fig:coppeliasim}A snapshot of the simulation environment
developed for the UMIRobot. The UMIRobot can interact with objects
and students can test their designs and control algorithms. \textbf{Simulation
file available at \protect\url{https://github.com/mmmarinho/umirobot}.}}
\end{figure}

One important take-away message of this course is that a roboticist
should always test in simulation first. To make that possible, a new
simulation environment (see Fig.~\ref{fig:coppeliasim}) was developed.
The simulation environment is openly available alongside the sample
programs.\footnote{\url{https://github.com/mmmarinho/umirobot}} The
simulation is based on a simulation platform free for educational
use (CoppeliaSim \cite{rohmer2013v}).

Using the GUI, students can perform teleoperation without the real
robot to test their designs, in a hassle-free simulation environment.
In addition, the users can use the GUI to only read signals from the
master device they developed, in order to test the motion of the robot
in the simulation. However, the complex interactions between objects
and the student's grippers for the teleoperation challenge is something
that, within the scope of this course, can only be achieved with the
real robot.

\subsection{Lectures}

The program designed for this course is composed of thirteen lectures,
spanning all topics related to robot teleoperation. Excluding the
guidance, introduction, and on-demand lectures, the main topics covered
are
\begin{itemize}
\item Robot simulation
\item Master interfaces
\item Introduction to CAD
\item Principles of electronics design
\item Arduino programming
\item Basics on servo motors
\end{itemize}
On-demand lectures have included, for instance, task-space controller
design and soldering. The contents depend on the background of the
students.

\subsection{Grading}

The grades are composed of an individual component and a group component.
The individual component is composed of the weekly attendance quiz
results. The quizzes usually have five multiple-choice questions whose
answers are given in that particular lecture. 

The teleoperation challenge was thought as a way to give the students
a clear goal. In the last lecture, the students present themselves,
the thought process behind their designs, the designs, and their results
performing the teleoperation challenge using videos. The evaluation
for the teleoperation challenge is made in four dimensions, namely
best gripper, best master device, smoothest teleoperation, and best
presentation. Each student confidentially scores the teams they do
not belong to. The team with highest average score is declared winner
for each individual dimension and those results are posted on the
course's website. Note that these results do not affect student grading.
Nonetheless, students are given a confidential space to express their
own contribution and to mention any possible issues they might have
had with teammates.

\section{Student distribution and feedback\protect\label{sec:Student-distribution-and}}

From the beginning of the course in summer 2021, it has been held
for a total of four semesters. A total of 73 kits have been prepared
and sent to students in eight countries, namely Japan (77\%), Australia
(10\%), China (7\%), Bangladesh, Indonesia, Singapore, India, and
Malaysia (with a combined 6\%).

\begin{figure}[t]
\begin{centering}
\includegraphics[width=0.75\linewidth]{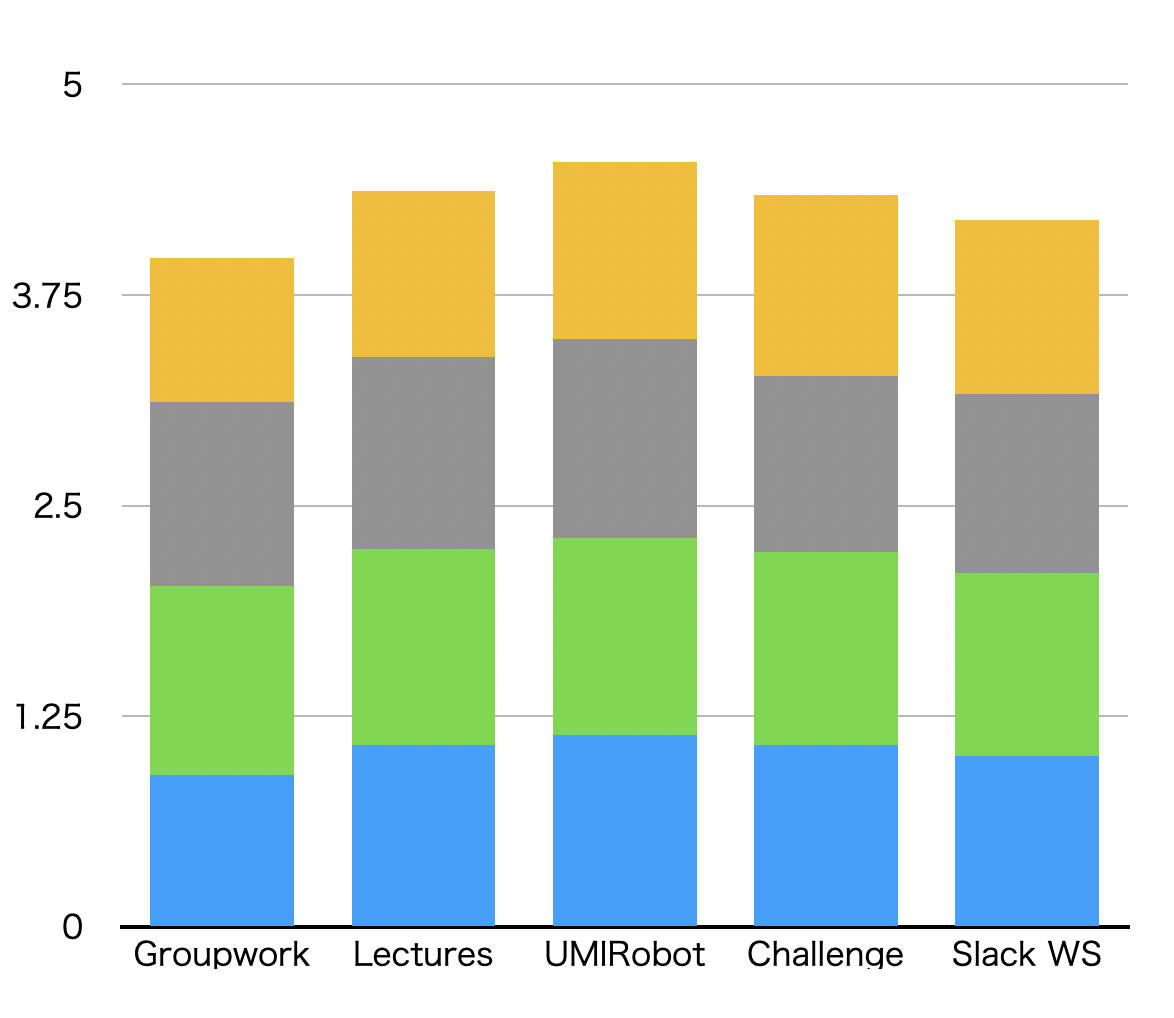}
\par\end{centering}
\caption{\protect\label{fig:course_rating}The average rating given by students
in all four semesters so far, where 5 is the highest possible value.
The colors encode the contribution to the average evaluation given
in each of the four semesters.}
\end{figure}

As part of the final lecture, the students also evaluate the course.
The students are asked to provide scores for relevant course components,
namely groupwork, lectures, UMIRobot, teleoperation challenge, and
slack workspace. The average rating given by students so far is summarized
in Fig.~\ref{fig:course_rating}. There is a visible trend that students
highly appreciate the UMIRobot, followed by the lectures. The groupwork
has the lowest score and it pertains the difficulties interacting
with students online as the majority of the students were unable to
meet in person. We believe that as the motion restrictions continue
to soften, we can move to a hybrid course in which our internationally-placed
attendees can enjoy the course online, whereas the university's regular
students can enjoy the activities in person.

\section{Achievement showcase\protect\label{sec:Achievement-showcase}}

\begin{figure*}[!t]
\includegraphics[width=1\textwidth]{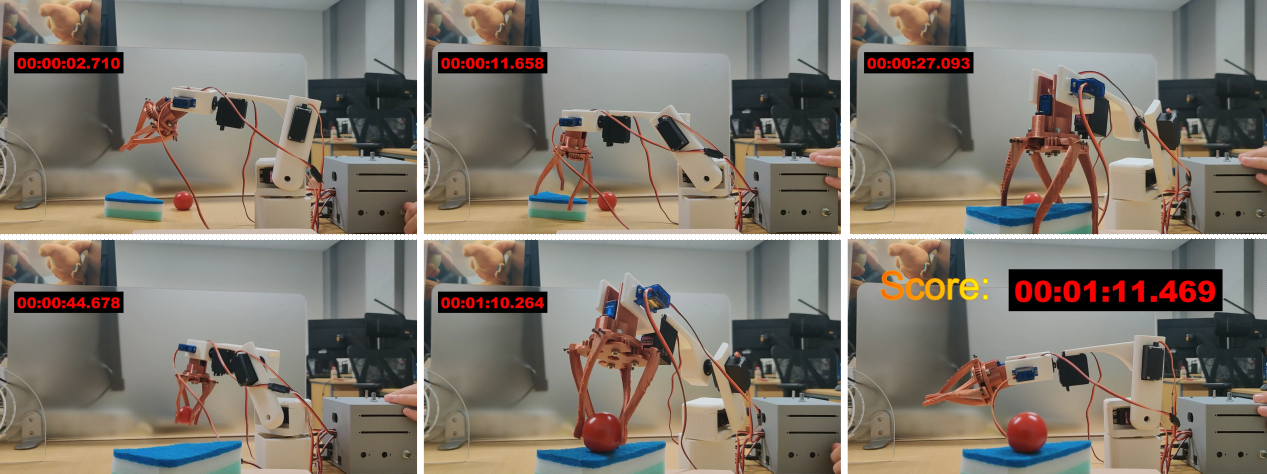}

\caption{\protect\label{fig:teleoperation_results}Snapshots of the teleoperation
challenge performed using the master device and gripper described
in Section~\ref{subsec:jzhao}. The user initially grabs the sponge,
which emulated a slice of cake. Then, the user places it in the correct
position. After, the user grabs the cherry tomato and places it on
top of the cake, finishing the task. The total time is shown in the
red timer, taking about 70s in total. \textbf{Video of this experiment
available at \protect\url{https://youtu.be/T65DRtAJ47Y}.}}
\end{figure*}

In this section we discuss two of the major achievements in the course
so far. First, two of the most creative gripper and master designs.
Second, an intercontinental teleoperation using the UMIRobot and additional
networking frameworks. There would be no enough room to discuss the
many great designs made in this course and many of those feature at
the course's website with the permission of their creators.

\textbf{These two results are shown in the attached video. Many other
videos showcasing the course's achievements are shown in \url{https://mmmarinho.github.io/UMIRobot/}.}

\subsection{Creative gripper and master device design\protect\label{subsec:jzhao}}

\begin{figure}[t]
\begin{centering}
\includegraphics[width=0.8\linewidth]{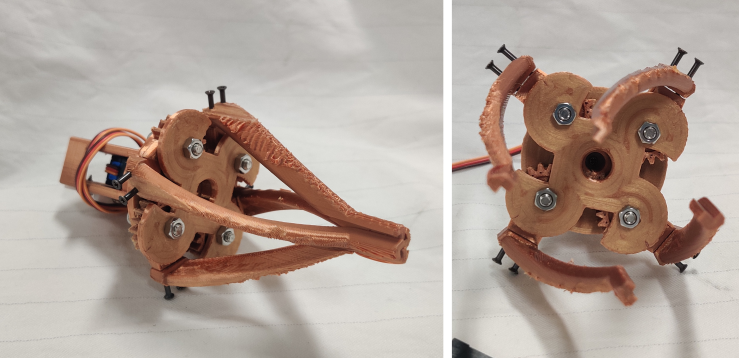}
\par\end{centering}
\caption{\protect\label{fig:jz_gripper}A gripper developed during the course.
The students are shown what was developed in prior courses and usually
feel motivated to try novel designs. Available at \protect\url{https://www.thingiverse.com/thing:6107830}.}
\end{figure}

\begin{figure}[t]
\begin{centering}
\includegraphics[width=0.75\linewidth]{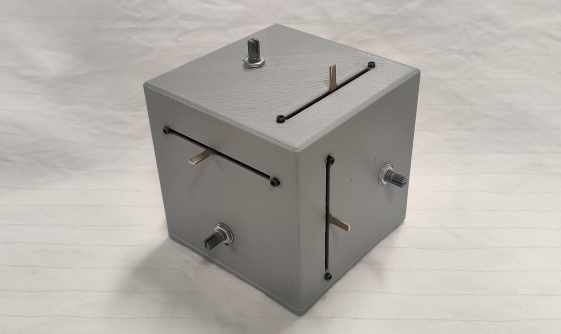}
\par\end{centering}
\caption{\protect\label{fig:jz_master}A master designed during the course.
It was designed to teleoperate the robot in each degree of freedom
of the task space. The linear potentiometers are used for linear motion
and the rotary potentiometers for rotational motion. By positioning
the master device and UMIRobot correctly, the user can operate using
an intuitive coordinate system. Available at \protect\url{https://www.thingiverse.com/thing:6107851}.}
\end{figure}

In the latest semester, the students decided on a task inspired on
placing a fruit at the top of a cake slice (emulated by a triangular
kitchen sponge). This task is quite challenging in that the fruit
is deformable and too much force can rip it apart. Conversely, not
enough force on it means it cannot be properly grasped.

In this context, one of the highlights of the course has been the
gripper and master shown in Figs. \ref{fig:jz_gripper} and \ref{fig:jz_master},
respectively. Whereas the students unfamiliar with robotics tend to
opt for the simplest designs, these two show a high level of creativity
and sense of mechanical design. It helps showcase that giving freedom
to students and a powerful technical support unleashes their creativity.

The gripper stands out for its complexity, having many gears and arms
with complicated shape. The design went through several revisions
to accommodate for the roughness of consumer-level 3D printers. The
master was designed to control the UMIRobot in task-space mode. Each
linear potentiometer was mapped into a translational degree of freedom,
whereas each rotational potentiometer was mapped to a rotational degree
of freedom. Despite their complexity, the master device and gripper
were effective in the teleoperation challenge as shown in Fig.~\ref{fig:teleoperation_results}.

\subsection{Intercontinental teleoperation\protect\label{subsec:quentin}}

\begin{figure}[t]
\begin{centering}
\includegraphics[width=0.9\linewidth]{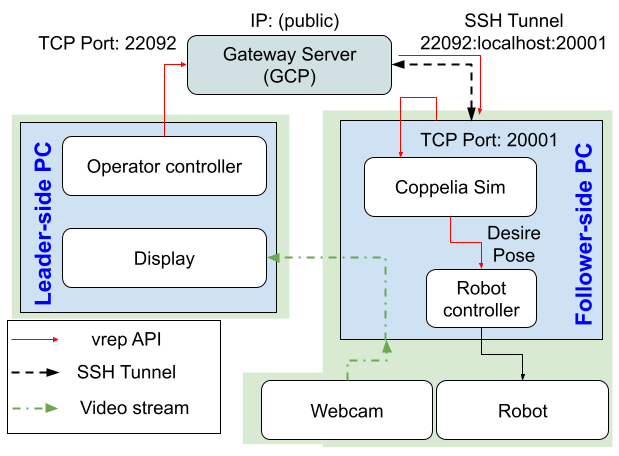}
\par\end{centering}
\caption{\protect\label{fig:ql_teleoperation}A block diagram of the most relevant
components used in the intercontinental teleoperation experiment.}
\end{figure}

For a course on robot teleoperation that can encompass undergrad students,
the teleoperation made by most students is mostly at the same desk,
or room. Nonetheless, the structure of the course allows some, with
more knowledge of computer networks to do more ambitious teleoperation
experiments. An example of this is the group in the first semester
with the highest know-how in networking which developed a framework
for intercontinental teleoperation based on CoppeliaSim, DQRobotics
\cite{Adorno2021}, and a secure shell (SSH) tunnel as summarized
in Fig.~\ref{fig:ql_teleoperation}.

\begin{figure}[t]
\begin{centering}
\includegraphics[width=0.9\columnwidth]{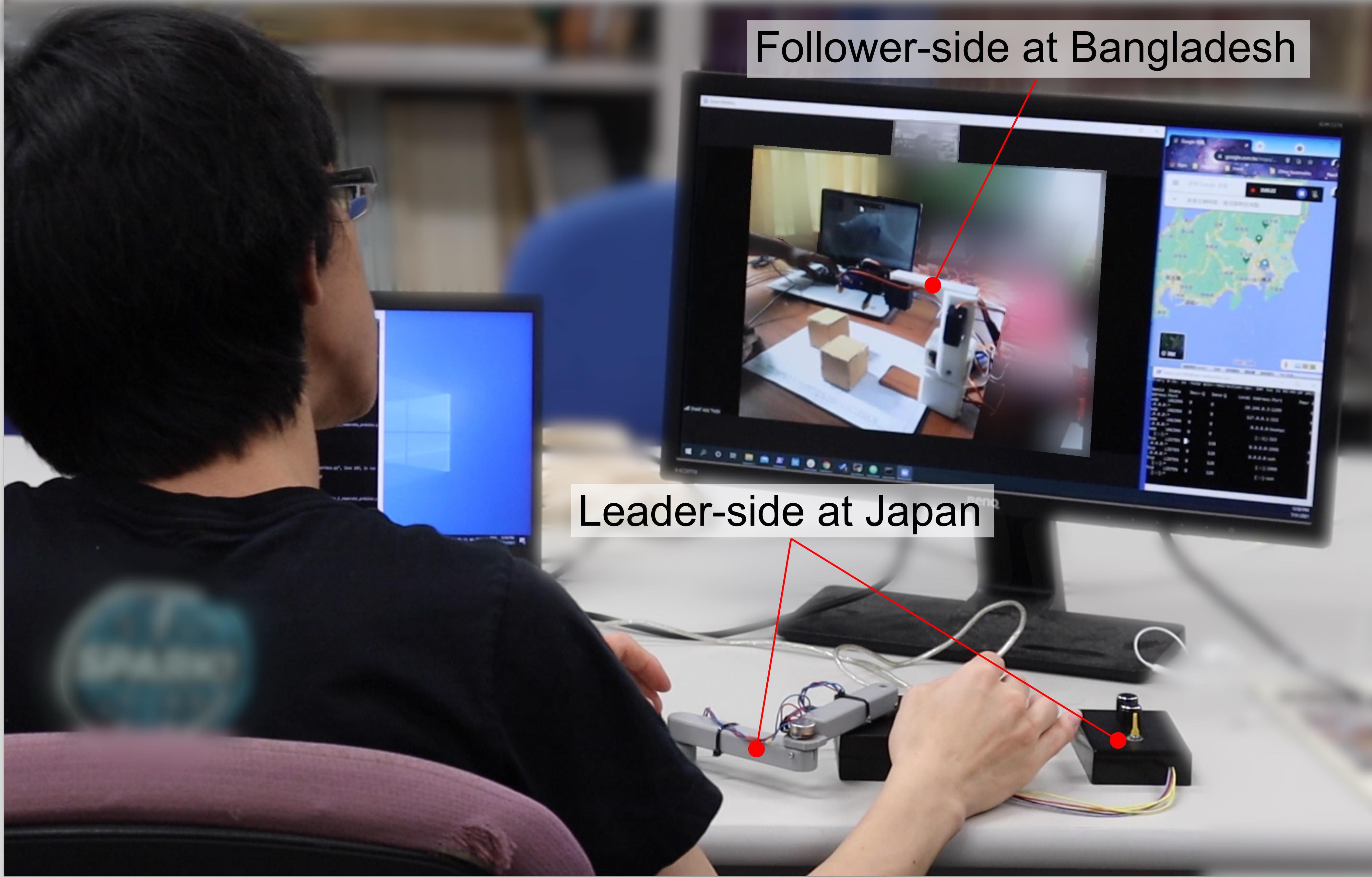}
\par\end{centering}
\cprotect\caption{A snapshot of the intercontinental teleoperation experiment. One student
in Japan controlled the master device designed by their group. The
UMIRobot in Bangladesh was controlled. The student could stack the
boxes naturally under teleoperation. \cprotect\textbf{Video of the
experiment available at \protect\url{https://youtu.be/Y_5amab3kMQ}.}}
\end{figure}

The intercontinental teleoperation experiment was performed across
Japan (leader-side) and Bangladesh (follower-side). A web conference
tool (Zoom, USA) and web camera were used for task visualization.
To send the commands to the robot, a public gateway for relay (Google
Cloud Platform Compute Instance) was used for a robust setup agnostic
of the network address translation settings on the ISPs of both sides.
During operation, the follower-side PC opened a TCP SSH forwarding
tunnel to the cloud gateway, and the leader-side PC was connected
directly to the follower's CoppeliaSim with provided gateway IP and
forwarded port.

Despite the presence of delay, as seen in the video in the supplementary
material, time delay was not an issue in this experimental showcase.
The kinematic control described in Section~\ref{subsec:Kinematic-control}
was effective enough.

\section{Conclusions}

In this work we presented the UMIRobot and its ecosystem for teleoperated
robots education. We described each element that compose the educational
kit and the course structure. We showed that students highly evaluate
the course and two examples of impactful results. First, the creative
design of a gripper and a master device, and a intercontinental teleoperation
experiment.

The proposed kit does not include a range of sensors (proximity, temperature,
etc) that are available in many hobbyist-oriented kit. Those are complementary
to the kit mentioned herein and can be used as sources for clever
and multi-modal control strategies. Differently from existing low-cost
manipulators available online from commercial sources, all parts of
the UMIRobot can be easily purchased or 3D printed given its open
hardware design. In addition, it has an open-source software infrastructure
allowing simple control strategies and state-of-the-art control.

The UMIRobot and the course will continue to evolve as society reopens
its borders. Nonetheless, the widespread push for online education
has left an infrastructural legacy that can be used to connect multiple
higher-education institutions across the world. There are ongoing
efforts in increasing the number of participating institutions and
to improve on online group work and other aspects to further cater
to student needs.

\bibliographystyle{IEEEtran}
\bibliography{bib/ral}

\end{document}